\documentclass[10pt,twocolumn,letterpaper]{article}
\usepackage[accsupp]{axessibility}
\usepackage{iccv}
\usepackage{times}
\usepackage{graphicx}
\usepackage{amsmath}
\usepackage{amssymb}

 \usepackage{subfigure} 
\usepackage{booktabs}
\usepackage{makecell}
\usepackage[ruled,linesnumbered]{algorithm2e}
\usepackage{authblk}

\usepackage[pagebackref=true,breaklinks=true,letterpaper=true,colorlinks,bookmarks=false]{hyperref}

\iccvfinalcopy % *** Uncomment this line for the final submission

\def\iccvPaperID{11470} % *** Enter the ICCV Paper ID here

\ificcvfinal\fi

\begin{document}

%%%%%%%%% TITLE
\title{Divide and  Conquer: 3D Point Cloud Instance Segmentation With Point-Wise Binarization}

\author{Weiguang Zhao\textsuperscript{1}, Yuyao Yan\textsuperscript{2}, Chaolong Yang\textsuperscript{1}, Jianan Ye\textsuperscript{2}, Xi Yang\textsuperscript{2}, Kaizhu Huang\textsuperscript{1}\thanks{Corresponding author}\\
\textsuperscript{1}Duke Kunshan University   $\qquad$  \textsuperscript{2}Xi'an Jiaotong-Liverpool University\\
{\tt\small \{weiguang.zhao, chaolong.yang, kaizhu.huang\}@dukekunshan.edu.cn}\\
{\tt\small \{jianan.ye20\}@student.xjtlu.edu.cn\quad \{yuyao.yan, xi.yang01\}@xjtlu.edu.cn}
}

\maketitle
\ificcvfinal\thispagestyle{empty}\fi

%%%%%%%%% ABSTRACT
\begin{abstract}
   Instance segmentation on point clouds is crucially important for 3D scene understanding.  Most SOTAs adopt distance clustering, which is typically effective but does not perform well in segmenting adjacent objects with the same semantic label (especially when they share neighboring points).  Due to the uneven distribution of offset points, these existing methods can hardly cluster all instance points.
   To this end, we design a novel divide-and-conquer strategy named PBNet that binarizes each point and clusters them separately to segment instances. Our binary clustering divides offset instance points into two categories: high and low density points (HPs vs. LPs). Adjacent objects can be clearly separated by removing LPs, and then be completed and refined by assigning LPs via a neighbor voting method. To suppress potential over-segmentation, we propose to construct local scenes with the weight mask for each instance. As a plug-in, the proposed binary clustering can replace the traditional distance clustering and lead to consistent performance gains on many mainstream baselines.  A series of  experiments on ScanNetV2 and S3DIS datasets indicate the superiority of our model. In particular, PBNet ranks first on the  ScanNetV2 official benchmark challenge, achieving the highest  $mAP$. Code will be available publicly at \href{https://github.com/weiguangzhao/PBNet}{\tt\small\text{https://github.com/weiguangzhao/PBNet}}.
\end{abstract}

%%%%%%%%% BODY TEXT
%=============================================================================Introduction=========================================================================================================
\section{Introduction}

In this paper, we consider instance segmentation for 3D point clouds that aims to classify each point of 3D clouds as well as separating objects from each class.
While a large body of successful algorithms have been developed for 2D images~\cite{ren2015faster,he2017mask,chen2019hybrid,liu2018path}, most of these methods are not particularly effective for 3D point clouds due to the inherent irregularity and sparsity in 3D data~\cite{dai2017bundlefusion,guo2020deep,yang2022towards}.

In 3D point cloud segmentation, PointGroup~\cite{jiang2020pointgroup} proposed a distance clustering framework to generate preliminary instance proposals. Although this framework is still being adopted by most SOTAs~\cite{he2021dyco3d,liu2021hida,chen2021hierarchical,vu2022softgroup,wu2022dknet}, it may usually have the following shortcomings: (1) distance clustering is limited to segment the adjacent objects with the same semantic label, especially when neighboring points are sticking together; (2) distance clustering only considers points within a distance threshold, which may generate incomplete instances. 

\begin{figure}[t]
  \centering
  \includegraphics[width=0.95\linewidth]{ 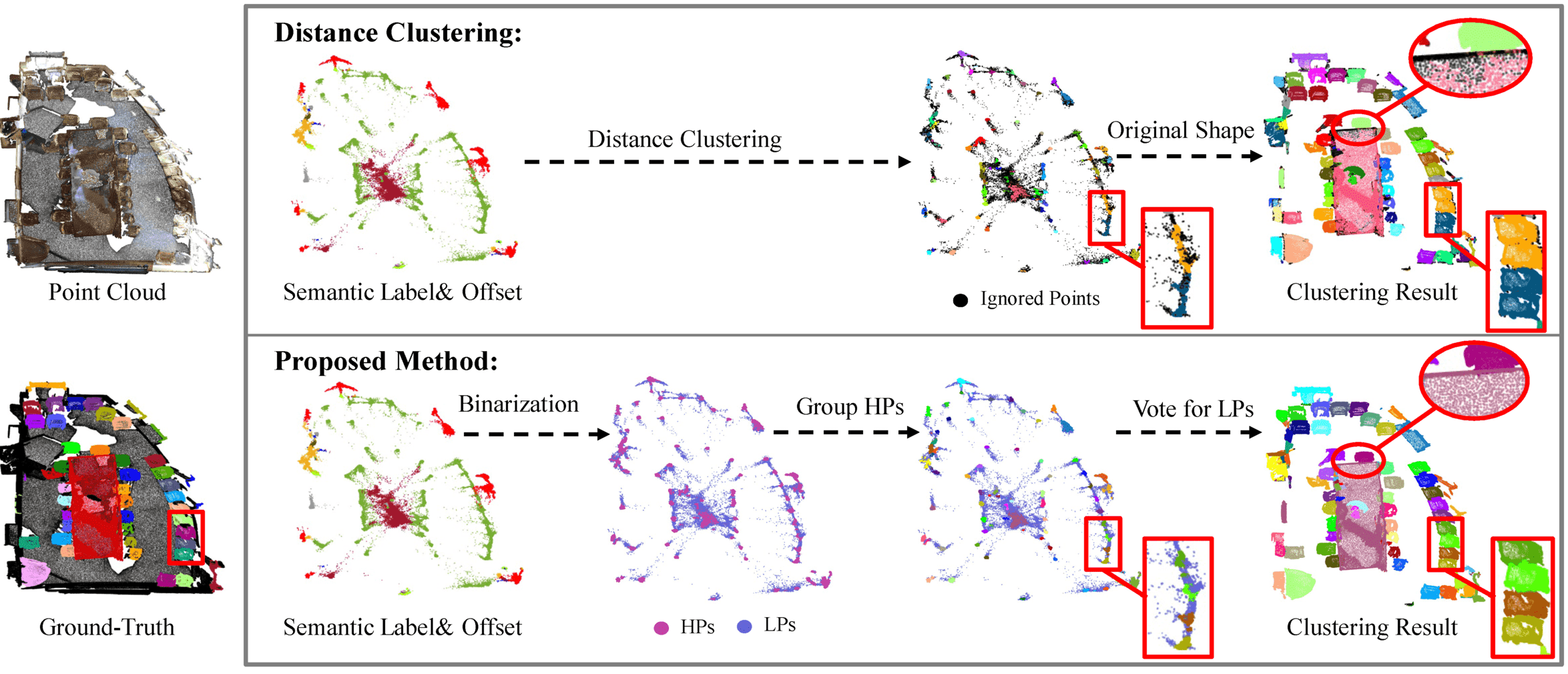}
   \caption{ Distance clustering vs. the proposed PBNet. Points dropped in the distance clustering are called ignored points. Clustering based on HPs can better segment adjacent instances ( highlighted in rectangular boxes), while the neighbor voting based on LPs can complete the instance (highlighted in ellipses).}
   \label{fig:intro_com}
\end{figure}

To alleviate these drawbacks, we propose a novel proposal generation framework to better segment adjacent objects and complete instances. Inspired by DBSCAN algorithm~\cite{ester1996density}, we divide foreground points into two categories: high and low density points (HPs vs. LPs), depending on the density of each point on the offset branch. As such, neighbor points between adjacent objects are binarized to LPs. Without the interference of neighbor points, grouping HPs can effectively separate adjacent objects. After that, we combine semantic prediction and neighbor voting to assign LPs. In this way, PBNet completely clusters all predicted instance points and works much more reasonable than the traditional distance clustering. The advantages of our methods are  illustrated in Fig.~\ref{fig:intro_com} where PBNet offers much better segmentation than the distance clustering. Notably, as shown in the experiments, by simply replacing the traditional distance clustering component,  the proposed binary clustering strategy  could also lead to significant performance gains on other mainstream baselines including PointGroup~\cite{jiang2020pointgroup} and HAIS~\cite{chen2021hierarchical}. 

Furthermore, taking into account the effects of offset error and density threshold, some larger objects such as sofas and tables have a certain probability of being divided into multiple instances. We further propose to search surrounding instances for each instance to construct the corresponding local scene. By designing a concise strategy, we encode each instance in each local scene to generate the corresponding weight mask, thus offering the network with prior knowledge to better focus  on the primary instance. Combining the global features and the local features, the final instance mask in the local scene will be predicted. Based on point-wise binarization and local scene, PBNet attains superior performance  on both ScanNetV2~\cite{dai2017scannet} and S3DIS~\cite{armeni20163d} dataset. The contributions of our work are as follows:

\begin{itemize}
	
	\item By dividing and conquering, we propose a novel clustering method based on binarized points to effectively segment adjacent objects and cluster all predicted instance points. It is appealing that by simply replacing the traditional distance clustering, our proposed  binary clustering strategy can also lead to significant performance gains on many mainstream baselines.
	\item We propose to construct local scenes combined with global feature and weight mask to refine instances, which can suppress over-segmentation and further boost the performance substantially.
	\item  Overall, we design a novel end-to-end 3D instance segmentation framework which significantly outperforms current SOTAs  for 3D instance segmentation: our model ranks the first on $mAP$  metric  of the ScanNetV2 official benchmark challenge.
	
\end{itemize}

%=============================================================================Related Work=========================================================================================================
\section{Related Work}

\subsection{Deep Learning on 3D Point Cloud}
PointNet~\cite{Qi_2017_CVPR} pioneered the application of deep learning techniques to point cloud processing.
Since then, deep learning has advanced significantly in a variety of 3D tasks, including 3D target detection, 3D semantic segmentation, 3D instance segmentation, 3D shape classification, and 3D reconstruction. Existing methods can be roughly divided into three categories: point-based, voxel-based, and multiview-based methods~\cite{guo2020deep}.
Point-based methods~\cite{qi2017pointnet++,wang2019dynamic,wu2019pointconv,rethage2018fully, wang2018sgpn} operate directly on the original points of the 3D point clouds without projection and volumetric operations.
Volumetric-based methods~\cite{maturana2015voxnet,riegler2017octnet} convert the 3D point clouds into a 3D volume representation and then extract features using a sparse convolution network.
Multiview-based methods~\cite{su2015multi,dai20183dmv,kundu2020virtual,hu-2021-bidirectional,jaderberg2015spatial} project 3D point clouds to multiple 2D planes in different directions to form multiple 2D images and then extract the features of these 2D images for feature fusion or analysis.

\subsection{Instance Segmentation for 3D Point Cloud}
Instance segmentation needs to separate each individual in the 3D scene, while semantic segmentation only needs to segment objects in the same category. The methods of 3D instance segmentation can be roughly divided into two categories: proposal-based and clustering-based. Proposal-based methods~\cite{hou20193d,qi2019deep,xie2020mlcvnet} are top-down approaches, which regress 3D bounding boxes to segment instances. GSPN~\cite{yi2019gspn} is an earlier proposal-based network. It abandons the traditional anchor-based method, and advocates learning what the target looks like before choosing the proposal region. 3D-BoNet~\cite{yang2019learning} develops a novel multi-criteria loss to constrain bounding boxes. 3D-MPA~\cite{engelmann20203d} combines a sparse convolutional network with a graph convolutional network to refine proposals.

Clustering-based methods dominate the benchmark challenge for this task, especially on ScanNetV2~\cite{dai2017scannet} dataset. These methods predict point-wise distance offsets from instance center points and group points on this branch. PointGroup~\cite{jiang2020pointgroup} takes point offset and distance clustering as the core of the algorithm. Many subsequent methods~\cite{he2021dyco3d,liu2021hida,chen2021hierarchical, vu2022softgroup, wu2022dknet} are all based on the distance clustering algorithm. HAIS~\cite{chen2021hierarchical} aggregates instances according to the number of points and designs the mask loss to refine instances. SSTNet~\cite{liang2021instance} utilizes superpoints to build a tree and aggregate the tree nodes to generate instances. SoftGroup~\cite{vu2022softgroup} adopts soft semantic predictions to reduce the impact of semantic error. DKNet~\cite{wu2022dknet} utilizes MLP to predict point-wise confidence based on distance clustering. DKNet can improve the segmentation of adjacent objects, but it also introduces confidence error and still ignores some foreground points. In contrast, PBNet binarizing points by point-wise density is more concise and effective.
\begin{figure*}[tp]
  \centering
  \includegraphics[width=0.95\linewidth]{ 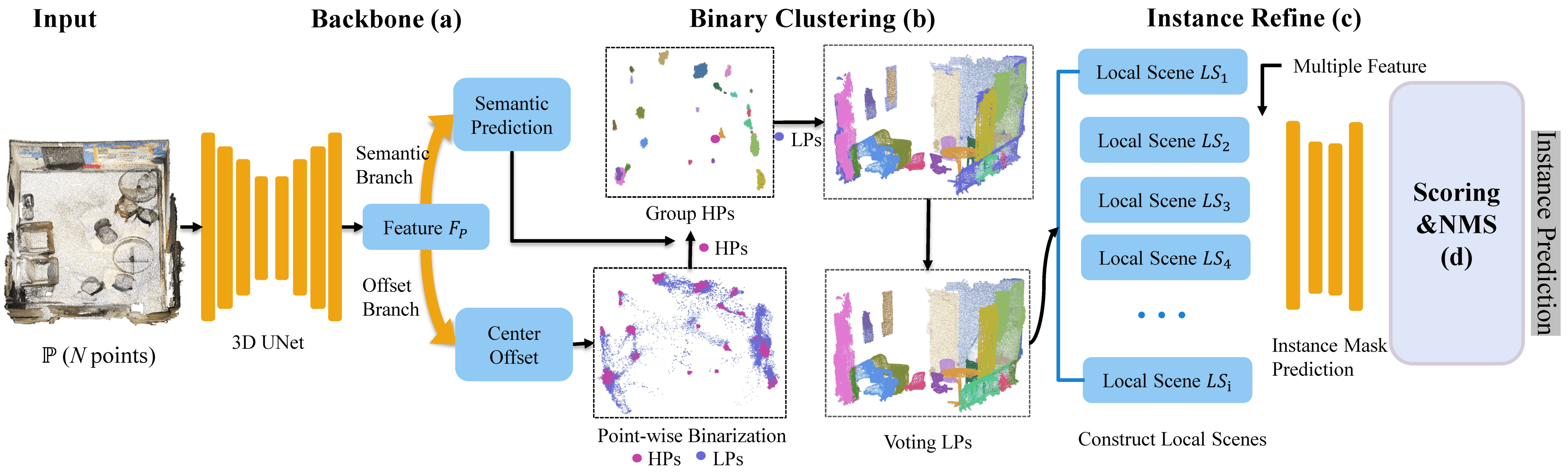}
   \caption{Network Architecture.}
   \label{fig:net}
\end{figure*}

Most SOTAs adopt merging-based method to suppress over-segmentation. HAIS~\cite{chen2021hierarchical} makes rules based on the average number of points contained in each category and average sizes of that to aggregate instances. MaskGroup~\cite{zhong2022maskgroup} sets an increasing distance threshold to merge instances iteratively. DKNet~\cite{wu2022dknet} learns the direct fusion relationship of each instance through the network to form a merging map, and utilize greedy algorithm to merge these instances. These methods are prone to under-segmentation while suppressing over-segmentation, and the instance edge cannot be refined by directly merging instances. Inspired by Knet~\cite{zhang2021k} and Mask-RCNN~\cite{he2017mask}, we construct the local scenes for each instance and generate the weight mask for local scene to implied different instances. Different from the existing SOTAs,  our methods is soft and combine global and local feature to refine instances. 

The proposed PBNet can be deemed as one voxel-based and clustering-based method. Different from the existing SOTAs, we divide the points into two categories  in the offset branch and process them separately. As shown in Fig.~\ref{fig:intro_com}, the adjacent objects with the same semantic label can be separated based on HPs. Meanwhile, handling LPs can better complete instances. Then PBNet construct the local scenes for each instance to suppress over-segmentation softly. PBNet demonstrate 
its superiority  to the other SOTAs.

%============================================================================Method===============================================================================================
\section{Our Method}

\subsection{Architecture Overview}
The overall network architecture of PBNet is depicted in Fig.~\ref{fig:net}. It consists of four main parts: Backbone~$\text{(a)}$, Binary Clustering~$\text{(b)}$, Instance Refine $\text{(c)}$, and Scoring \& NMS~$\text{(d)}$. First, traditional normal vectors are calculated on the faces~\footnote{The face is one base attribute of 3D items, often adopted in previous 3D instance segmentation works ~\cite{liang2021instance, wu2022dknet}.} of the point cloud.   We then feed the network with $xyz$, $rgb$, and normal vector features. 3D UNet~\cite{li2019gs3d,ronneberger2015u} and two FC layers are combined as a backbone to predict the point-wise semantic label and distance offset from the instance center. Then we calculate the density of each point on offset branch, and classify these points into two categories~(HPs/LPs) by setting the density threshold $\theta_d$. Combined with semantic prediction, HPs will be grouped to form preliminary instances. We convert the grouped HPs and ungrouped LPs in the offset coordinate system back to the original coordinate system. Furthermore, LPs will be assigned to the instances by neighbor voting algorithm. In order to suppress over-segmentation, we search surrounding instances for each instance to construct the corresponding local scene. The number of local scenes is the same as that of instances. Integrated with feature $F_p$, local scenes are utilized to refine each instance mask. Finally, we adopt ScoreNet~\cite{jiang2020pointgroup} and Non-maximum suppression (NMS) to achieve the instance prediction. 

\subsection{Backbone}
\label{RE_2_1}
Same as many SOTAs~\cite{jiang2020pointgroup,chen2021hierarchical,liang2021instance,vu2022softgroup,wu2022dknet}, 3D UNet~\cite{li2019gs3d,ronneberger2015u} is used to extract features of each point in our implementation. The point cloud is converted into a voxel form before it is fed into 3D UNet. When the features are extracted by 3D UNet, the voxel form point cloud is then converted to the point format according to the index. The semantic and offset branches composed of multi-layer perceptrons (MLP) are utilized to predict semantic label and offset for each point. At this stage, the background points (wall, floor) in the offset branch are removed according to the prediction of semantic results.

\noindent\textbf{Semantic Branch.}
The features of each point are fed into a 3-layer MLP to predict its semantic score of each class. The semantic scores are recorded as $\mathbf{S} \in [0,1]^{N \times M}$, where $N$ and $M$ are the number of point and class, respectively. The class with the highest score will be the semantic label for points. We utilize the cross-entropy loss $L_{sem}$ to regularize the semantic results.

\noindent\textbf{Offset Branch.} Similar to the semantic branch, we adopt a 3-layer MLP to predict offset vector $\boldsymbol{o}_{i} = \{o_x^i,o_y^i,o_z^i\}$ of each point, where $i\in \{1, \ldots, N\}$. Since $\boldsymbol{\hat{c}}_{i} = \{\hat{c}_x^i,\hat{c}_y^i,\hat{c}_z^i\}$ is the centroid of the instance that point $i$ belongs to, $L_{1}$ regression loss is taken to constrain points with the same instance labels to learn offsets ~\cite{qi2019deep,jiang2020pointgroup}. The calculation formula of $L_{1}$ regression loss is as follows:
\begin{equation}
	L_{ o_{-} dist}=\frac{1}{\sum_{i=1}^N} \sum_{i=1}^N\left\| \boldsymbol{o}_{i}-\left(\boldsymbol{\hat{c}}_{i}-\boldsymbol{p}_{i}\right)\right\| ,
\end{equation}
where $\boldsymbol{p}_{i}=\{p_x^{i}, p_y^{i}, p_z^{i}\}$ describes the 3D coordinate of point $i$ in the original point clouds. The calculation formula of $\boldsymbol{\hat{c}}_{i}$ is as follows: 

\begin{equation}
	\boldsymbol{\hat{c}}_{i}=\frac{1}{N_{map(i)}^{I}} \sum_{j \in I_{map(i)}} \boldsymbol{p}_{j} ,
\end{equation}
where $map(i)$ maps point $i$ to the index of its corresponding ground-truth instance. $N_{map(i)}^{I}$ is the number of points in instance $I_{map(i)}$. In order to regress the precise offsets, we follow~\cite{lahoud20193d,jiang2020pointgroup} to adopt direction loss $L_{o_{-} d i r}$:
\begin{equation}
	L_{o_{-} d i r}=-\frac{1}{\sum_{i=1}^N} \sum_{i=1}^N \frac{\triangle \boldsymbol{p}_{i}}{\left\|\triangle \boldsymbol{p}_{i}\right\|_{2}} \cdot \frac{\boldsymbol{\hat{c}}_{i}-\boldsymbol{p}_{i}}{\left\|\boldsymbol{\hat{c}}_{i}-\boldsymbol{p}_{i}\right\|_{2}}.
\end{equation}
This loss reinforces each point to move towards the correct direction by constraining the angle between the predicted offset vector and the ground-truth vector.

\subsection{Binary Clustering}

\begin{figure}[ht]
  \centering
  \subfigure[Point Density.]{\includegraphics[width=0.30\linewidth]{ 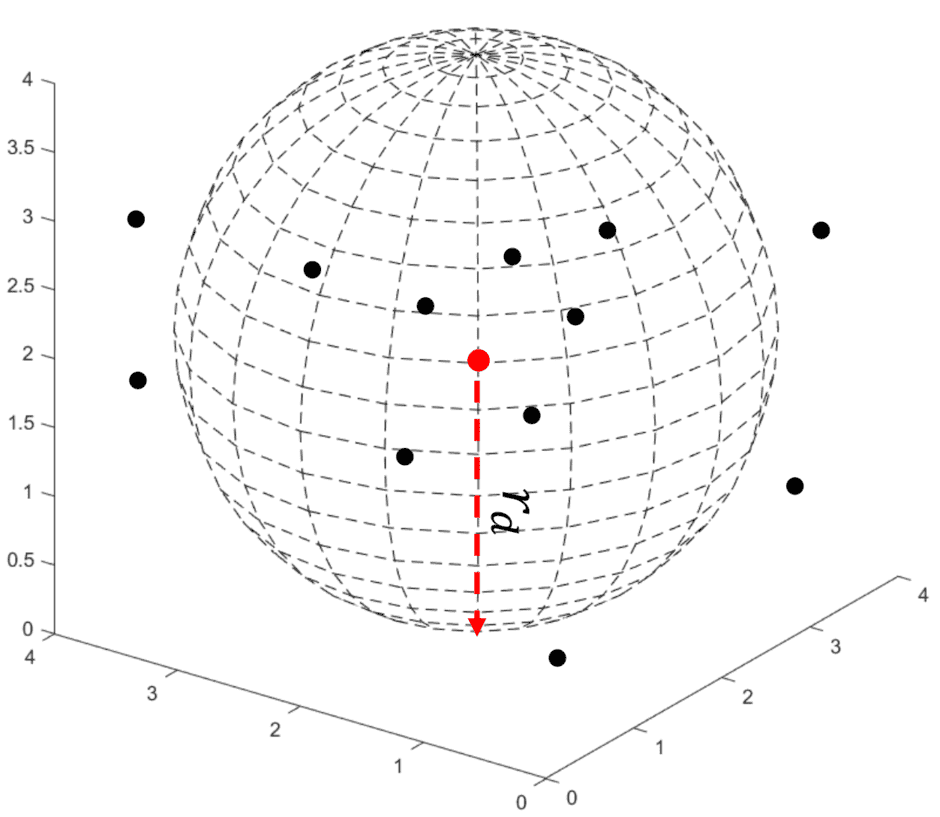}\label{fig:a}}
  \hfill
  \subfigure[Binarization.]{\includegraphics[width=0.65\linewidth]{ 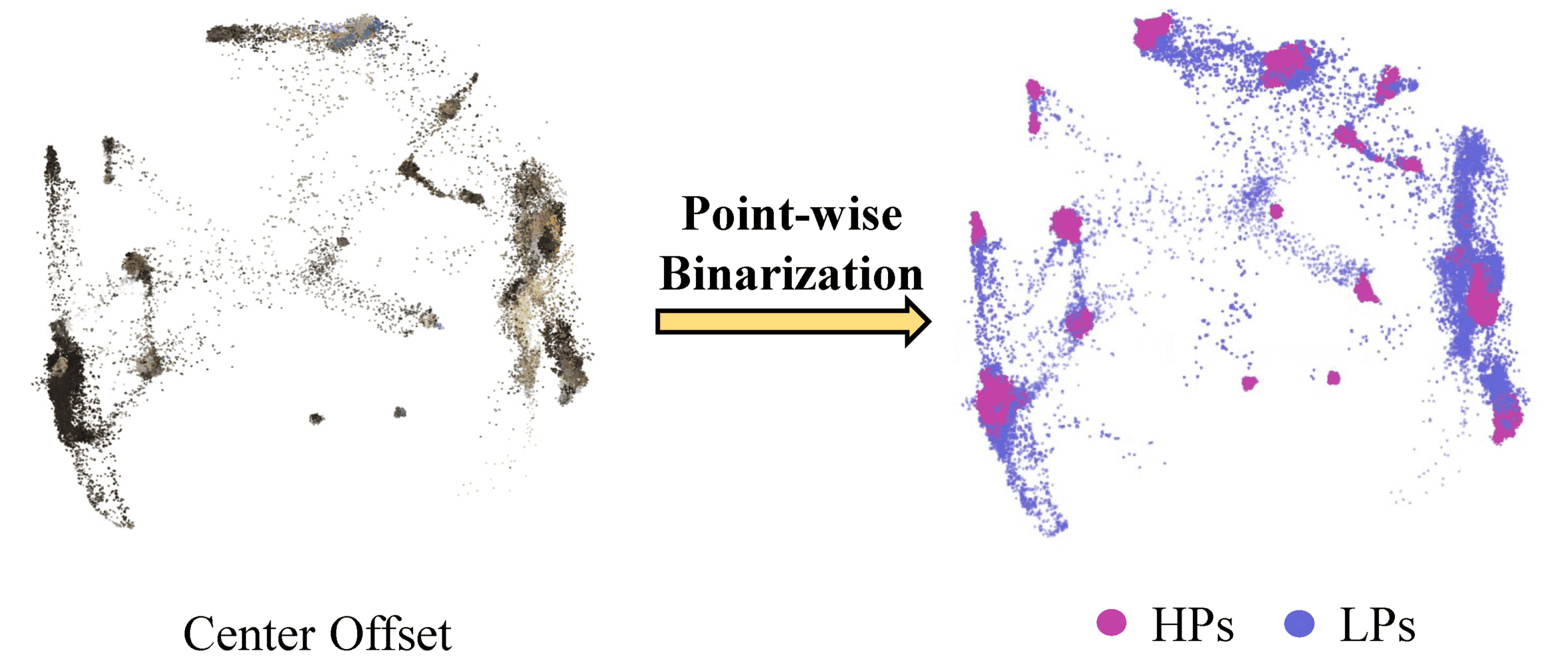}\label{fig:b}}
  \caption{Point-wise Binarization.}
  \label{fig:pb}
\end{figure}

\subsubsection{Point-wise Binarization}
We deploy point-wise density to conduct binarization. Its calculation process is shown in Fig.~\ref{fig:a}. For each point, we draw a sphere of radius $r_d$. The number of points in the ball is used to reflect  the density. Exactly, the density of  the point $p_i$  can be defined as the quantity of points within a sphere centred on the point $p_i$ with radius $r_d$. For example, in Fig.~\ref{fig:a}, the value reflecting the density of the red points is given as $7$. According to this method, we calculate the density value of every instance point on the offset branch. With the density of points, these points can be divided into two categories: HPs and LPs. If the densities of points are greater than the threshold $\theta_d$, these points are classified as HPs, while the remaining points will be classified as LPs.
\subsubsection{Grouping HPs}
We utilize semantic prediction and develop one modified variant of DBSCAN~\cite{ester1996density} to group  HPs. Specifically, we extend the traditional unsupervised DBSCAN 
to a weakly-supervised version by feeding semantic labels to guide clustering. With the weakly supervised information, PBNet can lead to much accurate clustering. Meanwhile, considering that the number of HPs is often huge, we further take binary search, and CUDA to speed up the clustering process. As a result, the time complexity can be substantially reduced from $\mathcal{O}\left(N_h^2\right)$ to $\mathcal{O}\left(N_h\log(N_h) / (K_c * T)\right)$, where $N_h$ is the number of HPs, $K_c$ is semantic category number, $T$ is thread number of CUDA. Overall, our HP grouping method is both accurate and fast.

\subsubsection{Voting LPs}
\begin{figure}[ht]
	%图片剧中
	\centering
	%设置图片大小、位置
	\includegraphics[width={0.95\linewidth}]{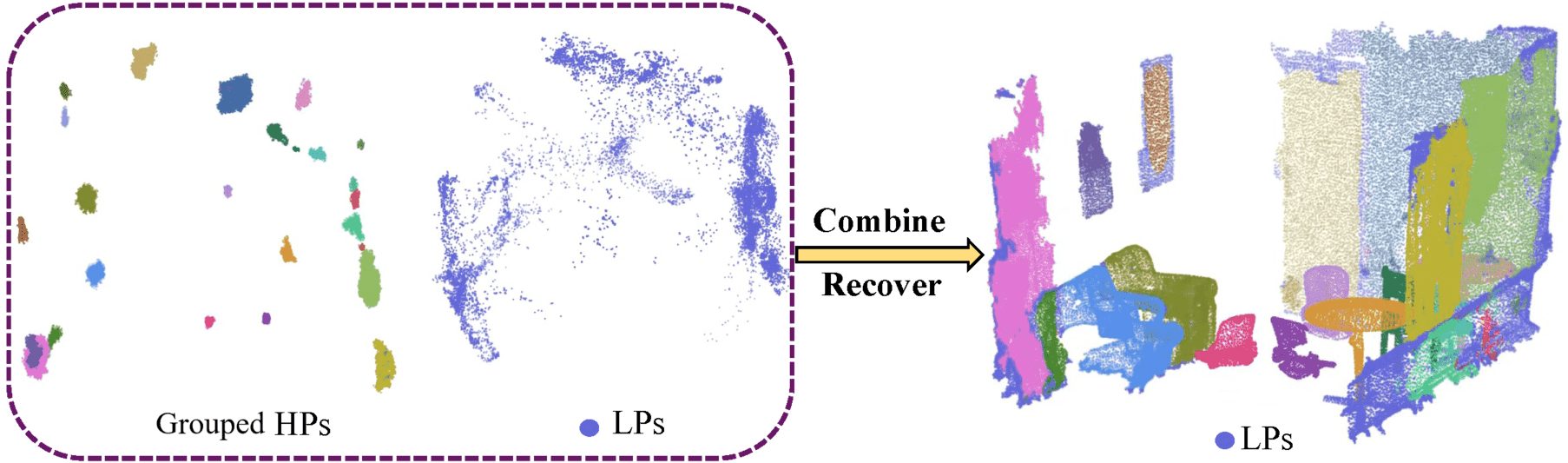}
	%设置图片lable、图片名称
	\caption{\label{fig:vote_com} Combination and Recovery}
\end{figure}

LPs are also critical to instance segmentation, which can lead to more complete and refined instances.  We combine LPs and grouped HPs, and change them back to the original shape according to the index. As shown in Fig.~\ref{fig:vote_com}, we find that all LPs are almost edge points. To this end, we develop neighbor voting~\cite{zhang2007ml} to determine which instance these LPs belong to. Different from the previous algorithm, we introduce the mean size of each category $r_{m}$ (which can be estimated from training data) and predict the semantic label to assist judgment. For each noise point, we select the HPs which share the same semantic label as LPs in the $r_{m}$ range. Then we count which instance these HPs belong to, and take the instance that contains the most HPs as the attribution of this noise point. There might be also an extreme case,  i.e., there are no HPs with the same semantics around the noise. In this case, we put aside the semantic label and directly exploit the nearest neighbor voting method~\cite{talavera2018big} to determine the attribution of the noise point. We repeat this operation until each noise point is classified. The time complexity of voting LPs is $\mathcal{O}\left(N_h*N_l/ (K_c * T)\right)$, where $N_l$ is the number of LPs.

\subsection{Instance Refining}

\subsubsection{Local Scene Construction}
\begin{figure}[t]
  \centering
  \subfigure[Construct Local Scene.]{\includegraphics[width=0.53\linewidth]{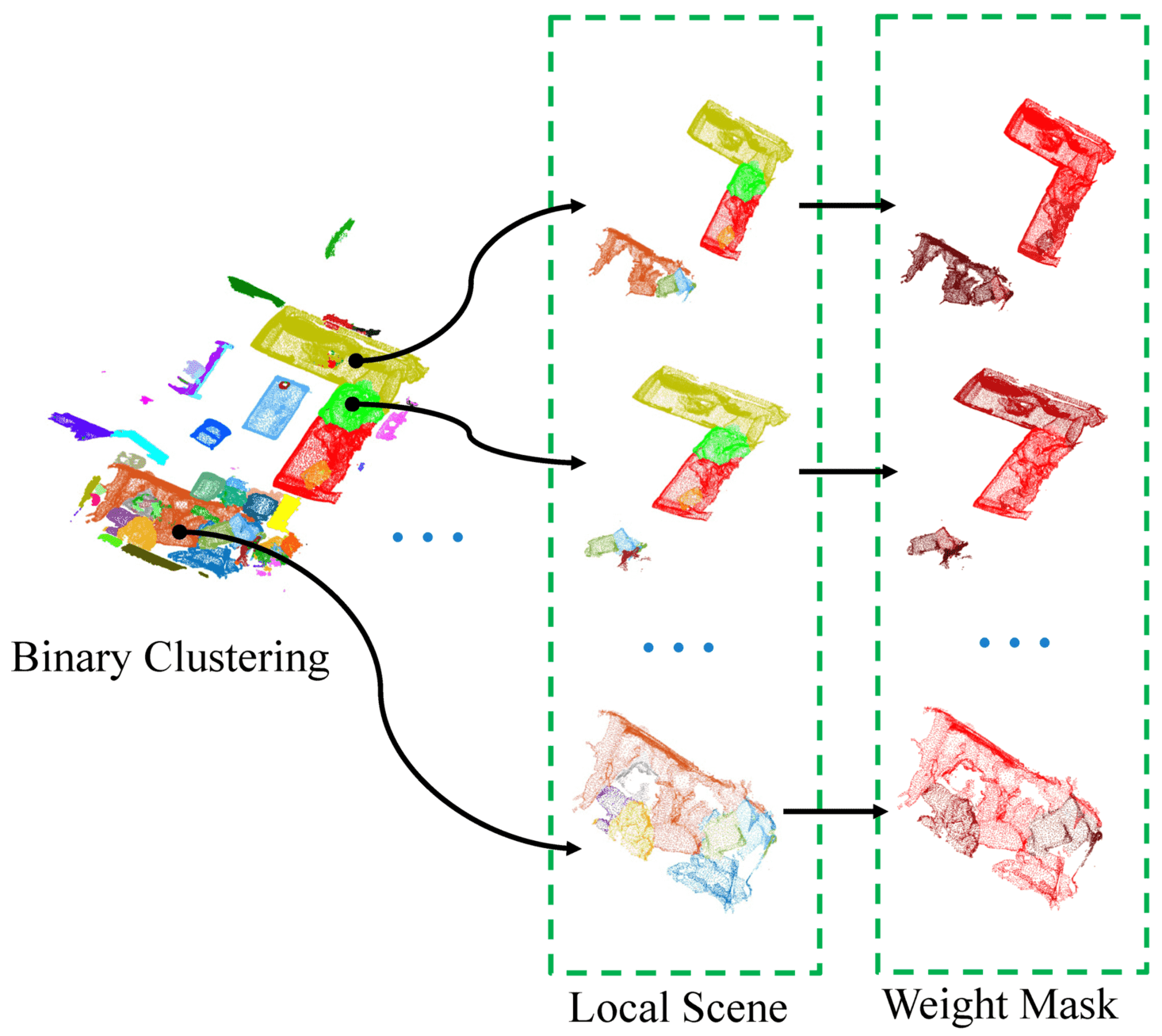}\label{fig:a1}}
  \hfill
  \subfigure[Instance Prediction.]{\includegraphics[width=0.42\linewidth]{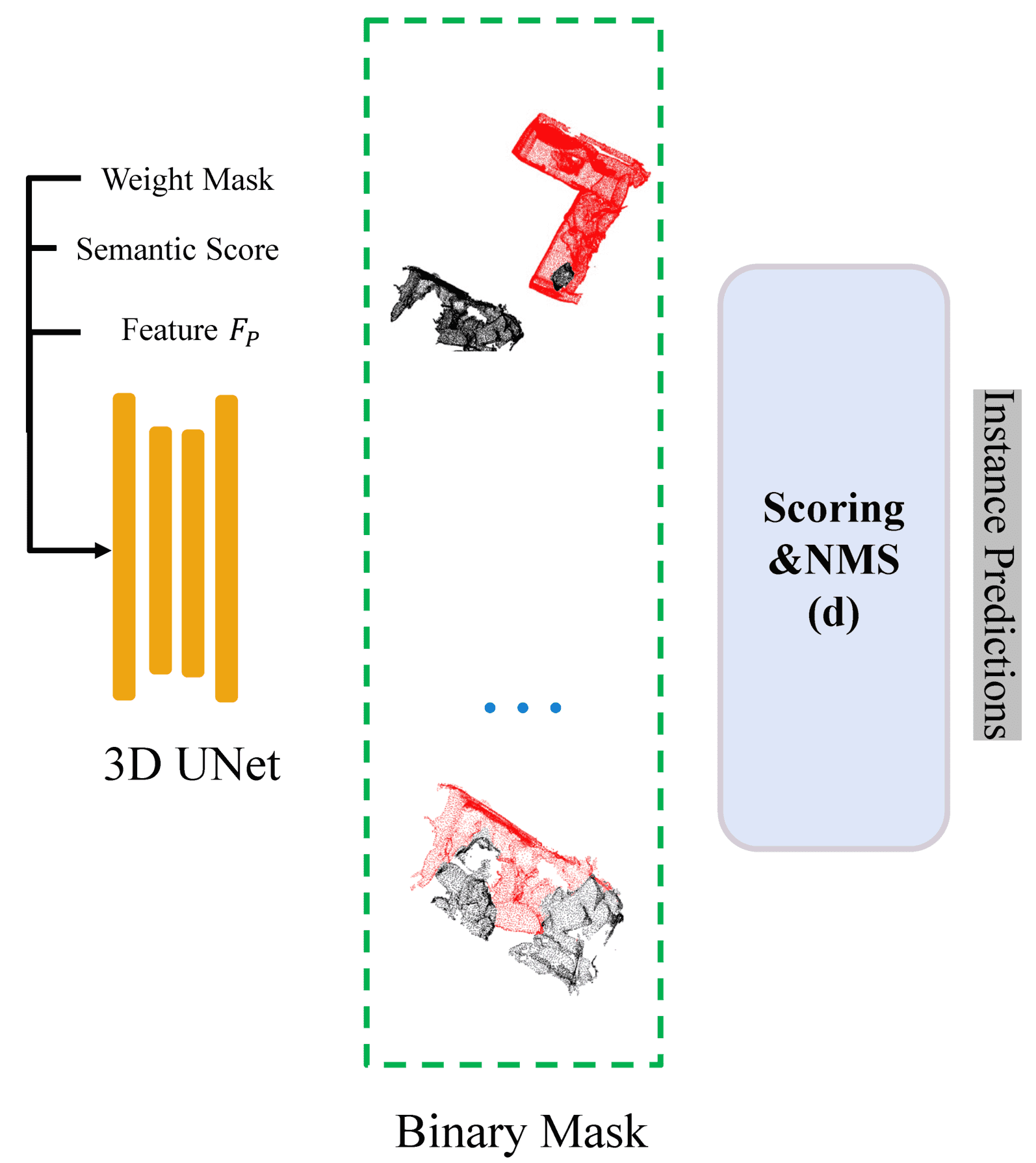}\label{fig:b1}}
  \caption{Instance Prediction.}
  \label{fig:local}
\end{figure}

Some objects with larger sizes and asymmetric shapes are easily over-segmented, such as the class of sofa as seen in  Fig.~\ref{fig:a1}. In the 2D domain, KNet~\cite{zhang2021k} proposes  that an object corresponds to an image mask. Due to the sheer size of the 3D scene, this method is difficult to be applied directly. Inspired by KNet, we propose to search the nearest $K$ instances (secondary instances) for each instance (primary instance). One local scene corresponds to one primary instance. To differentiate the primary and secondary instances in each scene, we define a concise formula to generate the weight mask. The calculation formula of weight masks $\mathbb{W}$ is as follows:
\begin{equation}
    \mathbb{W}_i = (Min(K, K_s-1) - i)/(Min(K, K_s-1)),
\end{equation}
where $\mathbb{W}_i$ is the weight mask of the $i$th closest secondary instance to the primary instance. $Min(\cdot)$ is the function that takes the minimum value. $K_s$ is the number of instances contained in the current semantic scene.

\subsubsection{Instance Mask Prediction}
Each local scene is fed to the 3D UNet to refine the mask of primary instances.  Weight mask, semantic score and feature $F_p$  are concatenated to be the input feature. Weight masks can provide prior knowledge to direct the network to focus more on the primary instance. The semantic score has been verified in the literature as one effective idea for instance segmentation~\cite{liang2021instance}. Feature $F_p$ is extracted by the backbone while the whole 3D scene is given as the input. The ground-truth mask of the local scene is defined as a binary mask, where ground-truth primary instance mask is 1 and others are 0.  Then we adopt the binary cross-entropy to calculate the mask prediction loss $L_{s_{-}mask}$:
\begin{equation}
	\begin{aligned}
		L_{s_{-}mask}=&-\frac{1}{\sum_{i=1}^{\hat{N}^{l}} \hat{N}_{i}} \cdot \sum_{i=1}^{\hat{N}^{l}} \sum_{j=1}^{\hat{N}_{i}}(y^{i}_{j} \log \left(\hat{y}^{i}_{j}\right) \\
		&+ \left(1-y^{i}_{j}\right)  \log \left(1-\hat{y}^{i}_{j}\right)),
	\end{aligned}
\end{equation}
where $\hat{N}^{l}$ is the number of local scenes. $\hat{N}_{i}$ denotes the number of points within the $i_{th}$ local scene and ${y}^{i}_{j}$ describes the ground-truth score of the $j_{th}$ points of the $i_{th}$ local scene. Moreover dice loss $L_{dice}$ is also applied, following DKNet~\cite{wu2022dknet}. The calculation formula of $L_{dice}$ is as follows:
\begin{equation}
L_{dice}=\frac{1}{\hat{N}^{l}} \sum_{i=1}^{\hat{N}^{l}}\left(\left(1-2 \frac{M^p_i \cdot M^g_i}{\left|M^p_i\right|+\left|M^g_i\right|}\right)\right),
\end{equation}
where $M^p_i$ and $M^g_i$ are predicted mask and ground-truth masks for the $i_{th}$ local scene, respectively.

\subsection{Scoring \& NMS}
Due to the over-segmentation, the primary instances may correspond to the same ground-truth instance after refinement. NMS is introduced to filter refined primary instances. Following~\cite{chen2021hierarchical}, we adopt ScoreNet~\cite{huang2019mask,jiang2018acquisition,li2019gs3d} to evaluate all instances and score them. ScoreNet consists of a lightweight 3D UNet and fully connected layers. For instance scores, we exploit a soft label $SC$ to supervise the predicted instance score $\hat{SC}$. Same as~\cite{jiang2020pointgroup}, the binary cross-entropy is used to calculate the instance score loss:

\begin{equation}
	\begin{aligned}
		L_{s_{-}ins}=&-\frac{1}{N_{ins}} \sum_{i=1}^{N_{ins}}({SC}_{i} \log \left(\hat{SC}_{i}\right)\\
		&+ \left(1-{SC}_{i}\right) \log \left(1-\hat{SC}_{i}\right)),
	\end{aligned}
\end{equation}
where $N_{ins}$ is the number of the predicted instances. We take the score as the confidence for each instance and utilize NMS to get the final instance result.

\subsection{Multi-Task Training}
\label{RE_2_6}
Our model can be trained in an end-to-end manner, even if it has multiple different tasks. The total loss of our network can be written as:
\begin{equation}
    \begin{aligned}
	{L}_{all}= &L_{sem} + L_{o_{-}dist} + L_{o_{-} d i r}  \\
	          &+ L_{s_{-}mask} + L_{dice}   + L_{s_{-}ins}
	\end{aligned}
\end{equation}

All loss weights are set to 1.0, which works well as empirically  verified in the experiments. Since $L_{s_{-}mask} , L_{dice} $ and $   L_{s_{-}ins}$ are affected by semantic and offset results, we do not add these losses until 128 epochs.

\section{Experiments}

\begin{table*}[t]
	%表格剧中
	\centering
	\resizebox{\textwidth}{!}{
	\begin{tabular}{c|c|cccccccccccccccccc}
	\bottomrule
	Method	&mAp & \rotatebox{90}{bathtub} & \rotatebox{90}{bed}   & \rotatebox{90}{bookshelf} & \rotatebox{90}{cabinet} & \rotatebox{90}{chair} & \rotatebox{90}{counter} & \rotatebox{90}{curtain} & \rotatebox{90}{desk}  & \rotatebox{90}{door}  & \rotatebox{90}{furniture} & \rotatebox{90}{picture} & \rotatebox{90}{refrigerato} & \rotatebox{90}{s.curtain} & \rotatebox{90}{sink}  & \rotatebox{90}{sofa}  & \rotatebox{90}{table} & \rotatebox{90}{toilet} & \rotatebox{90}{window} \\\hline
 		SGPN~\cite{wang2018sgpn}                        &4.9&2.3&13.4&3.1&1.3&14.4&0.6&0.8&0.0&2.8&1.7&0.3&0.9&0.0&2.1&12.2&9.5&17.5&5.4\\
		GSPN~\cite{yi2019gspn}                          &15.8&35.6&17.3&11.3&14.0&35.9&1.2&2.3&3.9&13.4&12.3&0.8&8.9&14.9&11.7&22.1&12.8&56.3&9.4\\
		3D-Bonet~\cite{yang2019learning}                &25.3&51.9&32.4&25.1&13.7&34.5&3.1&41.9&6.9&16.2&13.1&5.2&20.2&33.8&14.7&30.1&30.3&65.1&17.8\\
		3D-MPA~\cite{engelmann20203d}                   &35.5&45.7&48.4&29.9&27.7&59.1& 4.7&33.2&21.2&21.7&27.8&19.3&41.3&41.0& 19.5& 57.4& 35.2&84.9&21.3\\
		PointGroup~\cite{jiang2020pointgroup}           &40.7&63.9&49.6&41.5&24.3&64.5&2.1&57.0&11.4&21.1&35.9& 21.7&42.8&66.0& 25.6&56.2&34.1&86.0&29.1\\
		OCCuSeg~\cite{han2020occuseg}                   &48.6&80.2&53.6&42.8&36.9&70.2&20.5&33.1&30.1&37.9&47.4&32.7&43.7&\textbf{86.2}&48.5&60.1&39.4&84.6&27.3\\
		Dyco3d~\cite{he2021dyco3d}                      &39.5&64.2& 51.8&44.7&25.9&66.6&5.0&25.1&16.6&23.1&36.2&23.2&33.1&53.5&22.9&58.7&43.8&85.0&31.7\\
		PE~\cite{zhang2021point}                        &39.6&66.7&46.7&44.6&24.3&62.4&2.2&57.7&10.6&21.9&34.0&23.9&48.7&47.5&22.5&54.1&35.0&81.8&27.3\\
		SSTNet~\cite{liang2021instance}                 &50.6&73.8&54.9&49.7&31.6&69.3&17.8&37.7&19.8&33.0&46.3&57.6&51.5&85.7&\textbf{49.4}&63.7&45.7&94.3&29.0\\
		HAIS~\cite{chen2021hierarchical}                &45.7&70.4&56.1&45.7&36.4&67.3&4.6&54.7&19.4&30.8&42.6&28.8&45.4&71.1&26.2&56.3&43.4&88.9&34.4\\
		MaskGroup~\cite{zhong2022maskgroup}             &43.4&77.8&51.6&47.1&33.0&65.8&2.9&52.6&24.9&25.6&40.0&30.9&38.4&29.6&36.8&57.5&42.5&87.7&36.2\\
		SoftGroup~\cite{vu2022softgroup}                &50.4&66.7&57.9&37.2&38.1&69.4&7.2&\textbf{67.7}&30.3&38.7&\textbf{53.1}&31.9&58.2&75.4&31.8&64.3&49.2&90.7&\textbf{38.8}\\
        RPGN~\cite{dong2022learning}                    &42.8&63.0&50.8&36.7&24.9&65.8&1.6&67.3&13.1&23.4&38.3&27.0&43.4&74.8&27.4&60.9&40.6&84.2&26.7\\                                       
        PointInst3D~\cite{he2022pointinst3d}            &43.8&81.5&50.7&33.8&35.5&70.3&8.9&39.0&20.8&31.3&37.3&28.8&40.1&66.6&24.2&55.3&44.2&91.3&29.3\\
		DKNet~\cite{wu2022dknet}                        &53.2&81.5&\textbf{62.4}&51.7&37.7&\textbf{74.9}&10.7&50.9&30.4&43.7&47.5&\textbf{58.1}&53.9&77.5&33.9&64.0&50.6&90.1&38.5\\

        \textbf{Ours} &\textbf{57.3}&\textbf{92.6}&57.5&\textbf{61.9}&\textbf{47.2}&73.6&\textbf{23.9}&48.7&\textbf{38.3}&\textbf{45.9}&50.6&53.3&\textbf{58.5}&76.7&40.4&\textbf{71.7}& \textbf{55.9}&\textbf{96.9}&38.1 \\
	\bottomrule	
	\end{tabular}
	}
%设置表标题在下面
\caption{$mAP$ on ScanNetV2 Hidden Test Set.}

%设置label
\label{tab:com_t}
\end{table*}

\subsection{Experiment Setting}
\noindent\textbf{Datasets.} ScanNetV2~\cite{dai2017scannet}, one most challenging 3D dataset, includes 1,201 training samples, 312 validation samples, and 100 test samples where 20 semantic classes and 18 instance classes are labeled. Following most similar work in instance segmentation, classes including wall and floor are removed. The color for instance segmentation is random because the number of instances for each sample is flexible. We  compare the results on the validation as well as test set, which come from the official evaluation website.

S3DIS~\cite{armeni20163d} dataset includes 271 scenes within 6 areas. In these scenes, a total of 13 semantic classes are labeled. We utilize all the classes for instance evaluation and report the results on area 5, while the remaining areas are used for training. As points of the S3DIS scene is much more than ScanNetV2, we randomly sample points before each cropping by following the previous methods~\cite{jiang2020pointgroup,liang2021instance}.

\noindent\textbf{Evaluation Metric.} Following the ScanNetV2 official benchmark challenge, we report the mean average precision $AP$ ($mAP$) at overlap 0.25 ($AP_{25}$), overlap 0.5 ($AP_{50}$), and over overlaps in the range [0.5:0.95:0.05] ($AP$) for ScanNetV2 dataset.  Moreover, SoftGroup~\cite{vu2022softgroup} and DKNet~\cite{wu2022dknet}  also report the Box $AP_{50}$ and $AP_{25}$ results, which are commonly used in 3D object detection. For fair comparison, we follow them to report these metrics. Finally, we take the performance of $mAP$, $AP_{50}$, mean precision ($mPrec_{50}$) and mean recall ($mRec_{50}$) as the metric for S3DIS dataset, same as SOTAs. 

\noindent\textbf{Implementation Details.} We conduct training with two RTX3090 cards for 512 epochs. The batch size of training is set to 4. We adopt Adam~\cite{kingma2014adam} as the optimizer. The initial learning rate is set to 0.001 which decays with the cosine anneal schedule~\cite{loshchilov2016sgdr}. We set the voxel size to 0.02 by following pioneer methods~\cite{jiang2020pointgroup,chen2021hierarchical,he2021dyco3d}. For hyperparameters of density clustering, we tune  $r_d$, $\theta_d$ empirically as 0.04, 30 respectively.  The secondary instance number $K$ is empirically set to 7. The 3D UNet of backbone is MinkowskiNet34C~\cite{choy2019fully}, while the 3D UNet in mask prediction and ScoreNet are both MinkowskiNet14A~\cite{choy2019fully}. Data enhancements such as rotation, elastic distortion~\cite{ronneberger2015u}, color jittering, mixing~\cite{nekrasov2021mix3d} are adopted following the previous work~\cite{jiang2020pointgroup,nekrasov2021mix3d,vu2022softgroup}. Following SOTAs~\cite{wu2022dknet,han2020occuseg,liang2021instance}, a graph-based post-processing is utilized to smooth labels.

%%========================================
\begin{table}[b]
	%调整行距
	\setlength{\tabcolsep}{6pt}
	% 	\renewcommand{\arraystretch}{1.3}
	%表格剧中
	\centering
	\resizebox{0.45\textwidth}{!}{
		\begin{tabular}{c|ccc|cc}
			\hline
			& \multicolumn{3}{c}{Segmentation}                                            & \multicolumn{2}{c}{Detection}                                                       \\\hline
			                                       & $mAP$& $AP_{50}$ & $AP_{25}$      & Box $AP_{50}$& Box $AP_{25}$                        \\
			VoteNet~\cite{qi2019deep}               &-&-&-                 &33.5&58.6 \\
			3D-MPA~\cite{engelmann20203d}           &35.3 & 59.1 & 72.4    &49.2&64.2 \\
			PointGroup~\cite{jiang2020pointgroup}   &34.8& 56.9&71.3       &48.9&61.5  \\
			Dyco3D~\cite{he2021dyco3d}              &40.6& 61.0& -         &39.5&64.1 \\
			HAIS~\cite{chen2021hierarchical}       &43.5&64.1&75.6         &53.1&64.3 \\
			SSTNet~\cite{liang2021instance}        &50.0&64.7&73.9         &52.7&62.5 \\
			SoftGroup~\cite{vu2022softgroup}       &46.0&67.6&\textbf{78.9}         &59.4&\textbf{71.6} \\
                PointInst3D~\cite{he2022pointinst3d}   &45.6 & 63.7&-           &51.0&-  \\
			DKNet~\cite{wu2022dknet}               &51.5&67.0&77.0         &59.0&67.4 \\
			\textbf{Ours}                          &\textbf{54.3}&\textbf{70.5}&\textbf{78.9} &\textbf{60.1} & 69.3\\\hline         
		\end{tabular}
	}
	%设置表标题在下面
	\caption{Quantitative Comparison on ScanNetV2 Validation Set.}
	%设置label
	\label{tab:com_v}
\end{table}

\subsection{Comparison to SOTAs}

\begin{table}[h]
	%调整行距
	\setlength{\tabcolsep}{6pt}
	% 	\renewcommand{\arraystretch}{1.3}
	%表格剧中
	\centering
		\resizebox{0.40\textwidth}{!}{
	\begin{tabular}{c|cccc}
		\hline
		                                       & $mAP$  &$AP_{50}$ & $mPrec_{50}$ & $mRec_{50}$ \\ \hline
		SGPN$^{\dagger}$~\cite{wang2018sgpn}               &-&-&36.0&28.7	\\
		Dyco3D$^{\dagger}$~\cite{he2021dyco3d}             &-&-&64.3&64.2\\
		PointGroup$^{\dagger}$~\cite{jiang2020pointgroup}  &-&57.8&61.9&62.1  \\
		HAIS$^{\dagger}$~\cite{chen2021hierarchical}       &-&-&71.1 &65.0  \\
		SSTNet$^{\dagger}$~\cite{liang2021instance}        &42.7&59.3&65.5&64.2 \\
		MaskGroup$^{\dagger}$~\cite{zhong2022maskgroup}    &-&65.0&62.9 &64.7  \\
		SoftGroup$^{\dagger}$~\cite{vu2022softgroup}       &51.6& 66.1& 73.6& \textbf{66.6}\\
        RPGN$^{\dagger}$~\cite{dong2022learning}                       &-&-&64.0&63.0 \\
        PointInst3D$^{\dagger}$~\cite{he2022pointinst3d}               &-&-&73.1&65.2\\ 
		DKNet$^{\dagger}$~\cite{wu2022dknet}               &-& -& 70.8& 65.3\\
		\textbf{Ours}$^{\dagger}$                          &\textbf{53.5}&\textbf{66.4} & \textbf{74.9}   & 65.4\\ \hline
  
            SGPN$^{\ddagger}$~\cite{wang2018sgpn}               &-&-&38.2&31.2	\\
		PointGroup$^{\ddagger}$~\cite{jiang2020pointgroup}  &-&64.0&69.6&69.2  \\
		HAIS$^{\ddagger}$~\cite{chen2021hierarchical}       &-&-&73.2 &69.4  \\
		SSTNet$^{\ddagger}$~\cite{liang2021instance}        &54.1&67.8&73.5&73.4 \\
		MaskGroup$^{\ddagger}$~\cite{zhong2022maskgroup}    &-&69.9&66.6 &69.6  \\
		SoftGroup$^{\ddagger}$~\cite{vu2022softgroup}       &54.4& 68.9& 75.3& 69.8\\
        RPGN$^{\ddagger}$~\cite{dong2022learning}                        &-&-&\textbf{84.5}&70.5\\
        PointInst3D$^{\ddagger}$~\cite{he2022pointinst3d}                &-&-&76.4&\textbf{74.0}\\
		DKNet$^{\ddagger}$~\cite{wu2022dknet}               &-& -& 75.3& 71.1\\
            \textbf{Ours}$^{\ddagger}$                          &\textbf{59.5}& \textbf{70.6}& 80.1&72.9\\ \hline
            
	\end{tabular}
	}
	%设置label
	\label{tab:com}
	%设置表标题在下面
	\caption{Quantitative Comparison on S3DIS. $\dagger$ and $\ddagger$ indicate respectively the results  on Area 5 and 6-fold cross-validation.}
	%设置label
	\label{tab:res_s3dis}
\end{table}

\begin{figure*}[h]
	\centering 
	\begin{minipage}{0.18\linewidth}
		\centerline{\includegraphics[width=\textwidth]{ 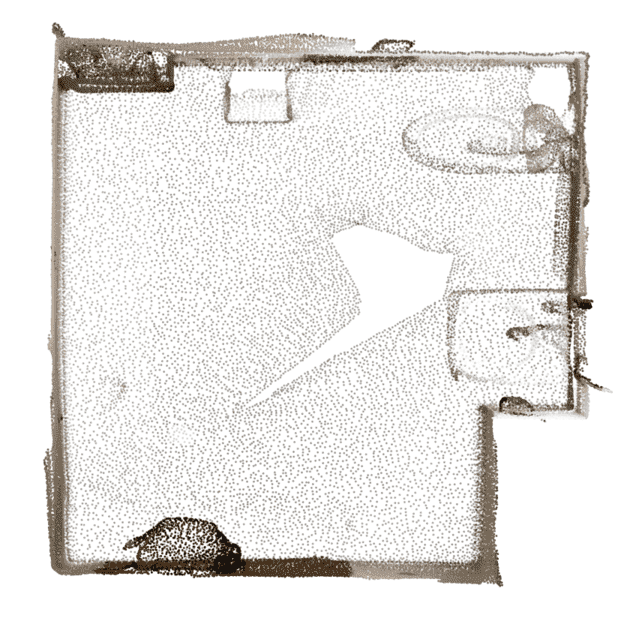}}
		\centerline{\includegraphics[width=\textwidth]{ 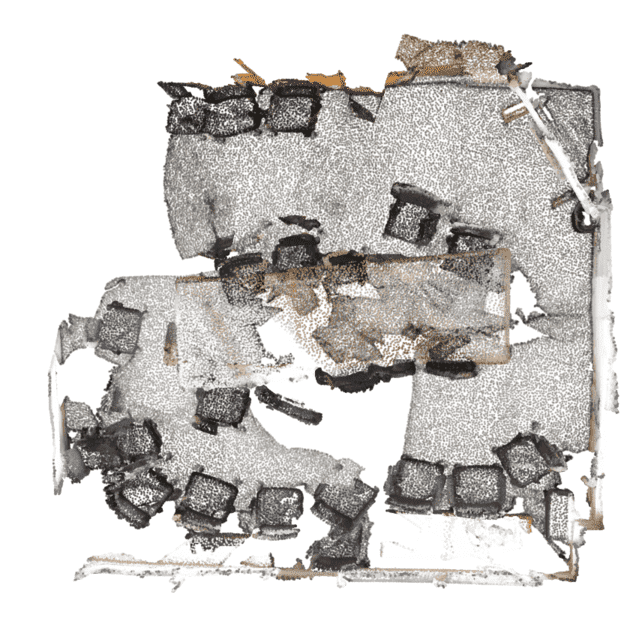}} 
		\centerline{\includegraphics[width=\textwidth]{ 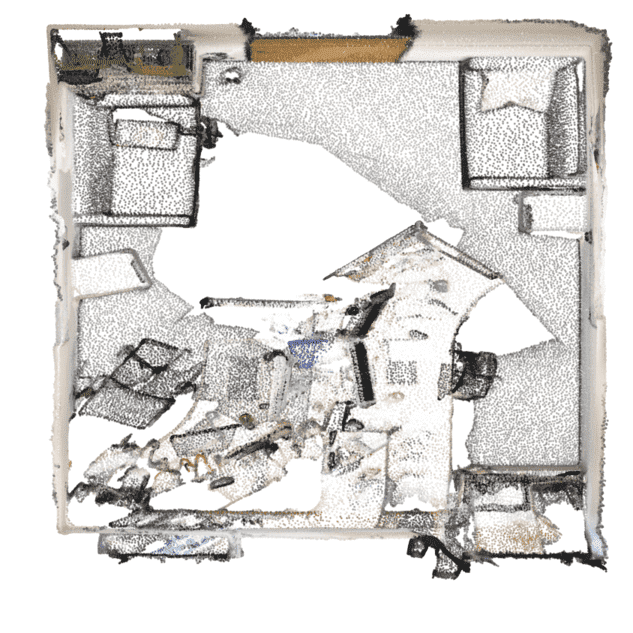}}
		\centerline{Input}
	\end{minipage}
	\begin{minipage}{0.18\linewidth} 
		\centerline{\includegraphics[width=\textwidth]{ 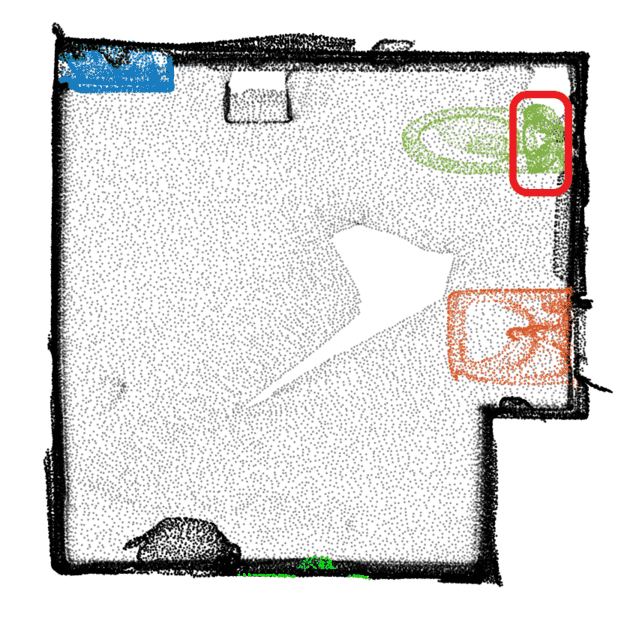}}
		\centerline{\includegraphics[width=\textwidth]{ 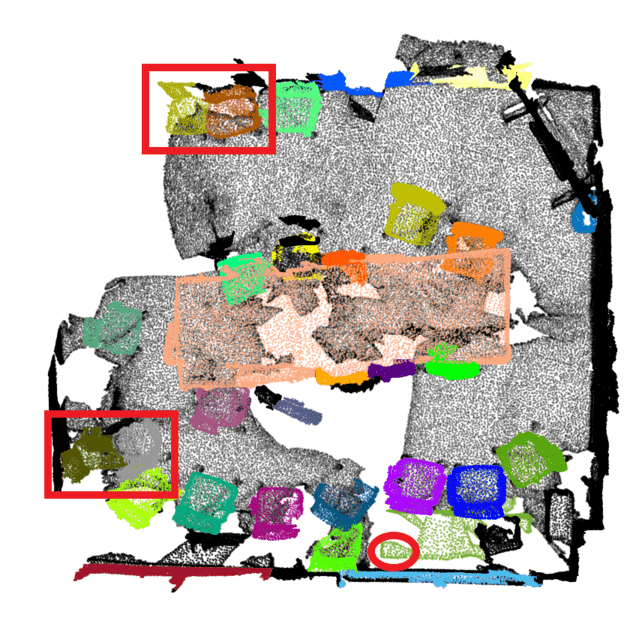}}
		\centerline{\includegraphics[width=\textwidth]{ 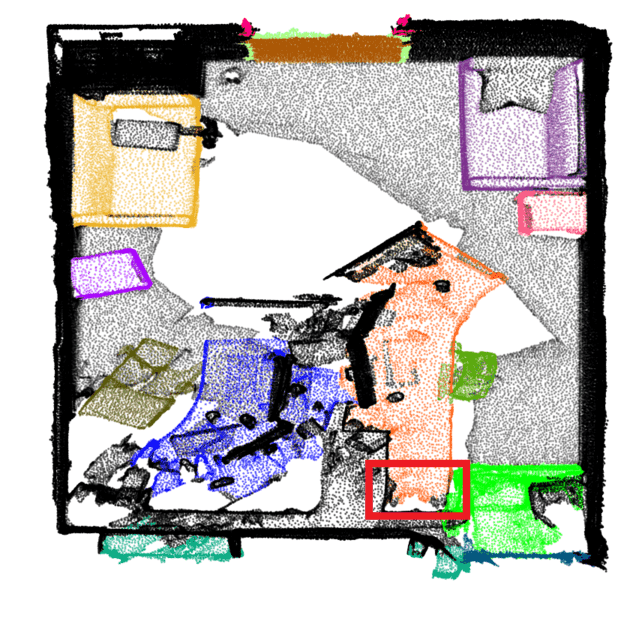}}
		\centerline{Instance\quad GT}
	\end{minipage}
	\begin{minipage}{0.18\linewidth}
		%\vspace{3pt}  是调图片之间的间隔  
		\centerline{\includegraphics[width=\textwidth]{ 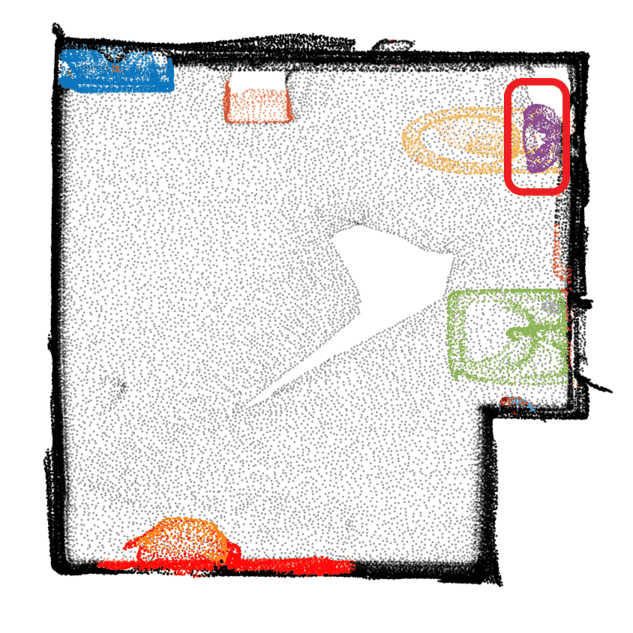}}
		\centerline{\includegraphics[width=\textwidth]{ 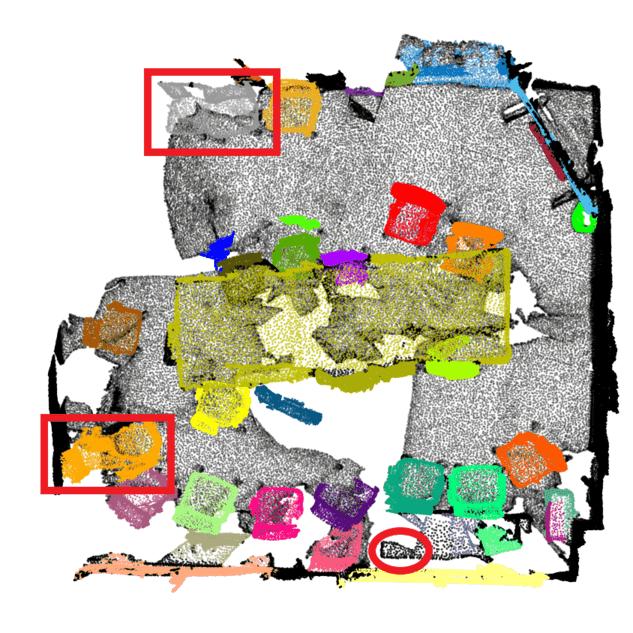}}
		\centerline{\includegraphics[width=\textwidth]{ 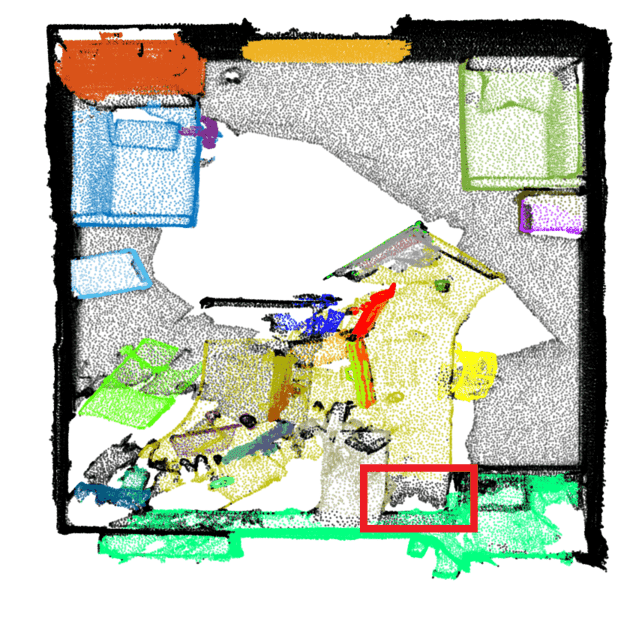}}
		\centerline{PointGroup~\cite{jiang2020pointgroup} }
	\end{minipage}
	\begin{minipage}{0.18\linewidth}
		%\vspace{3pt}  是调图片之间的间隔 
		\centerline{\includegraphics[width=\textwidth]{ 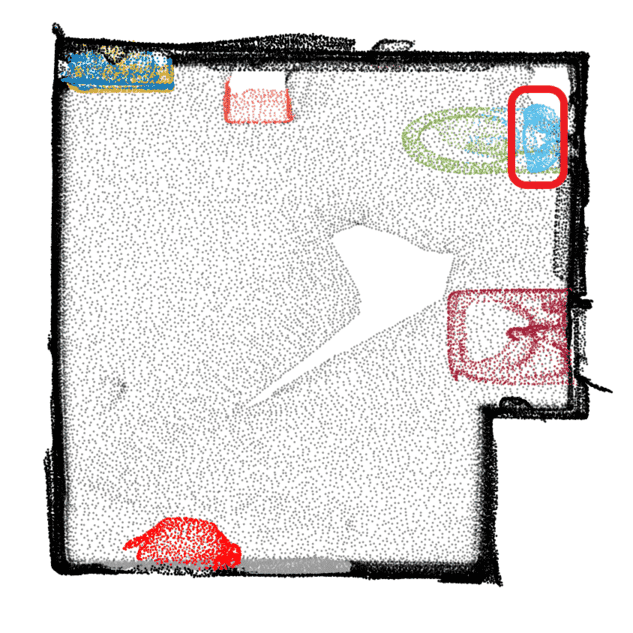}}
		\centerline{\includegraphics[width=\textwidth]{ 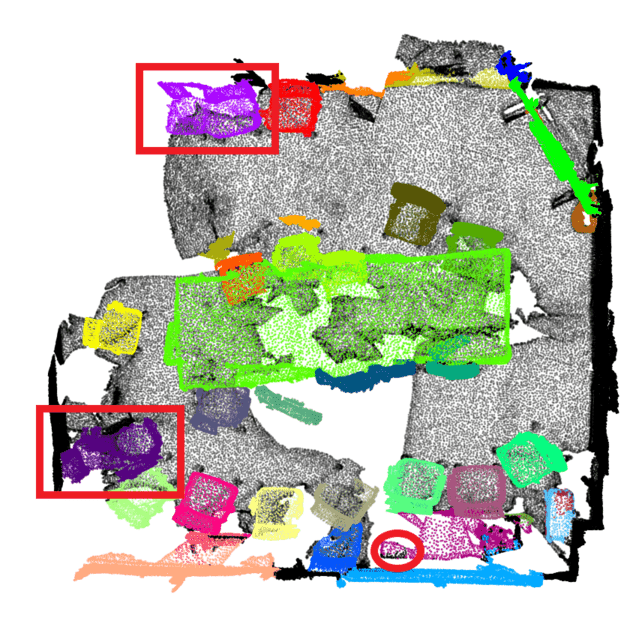}}
		\centerline{\includegraphics[width=\textwidth]{ pic_compress/compare/scene0025_01_ins_pred_hais.png}}
		\centerline{HAIS~\cite{chen2021hierarchical} }
	\end{minipage}
	\begin{minipage}{0.18\linewidth}
		%\vspace{3pt}  是调图片之间的间隔 
		\centerline{\includegraphics[width=\textwidth]{ 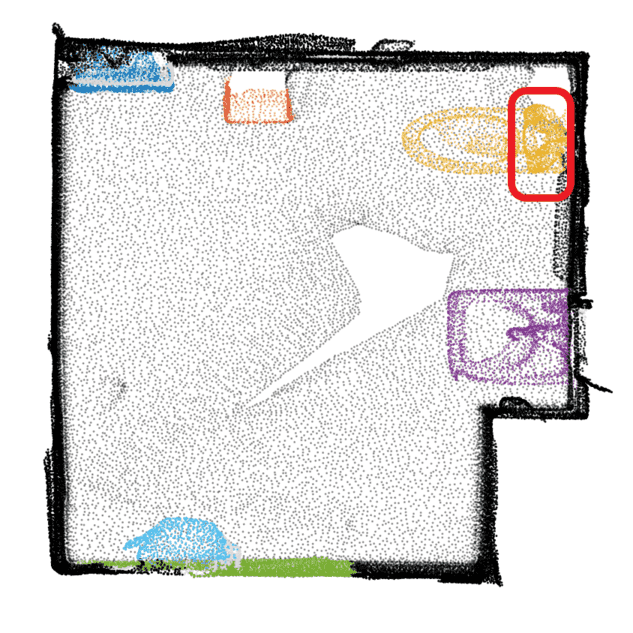}}
		\centerline{\includegraphics[width=\textwidth]{ 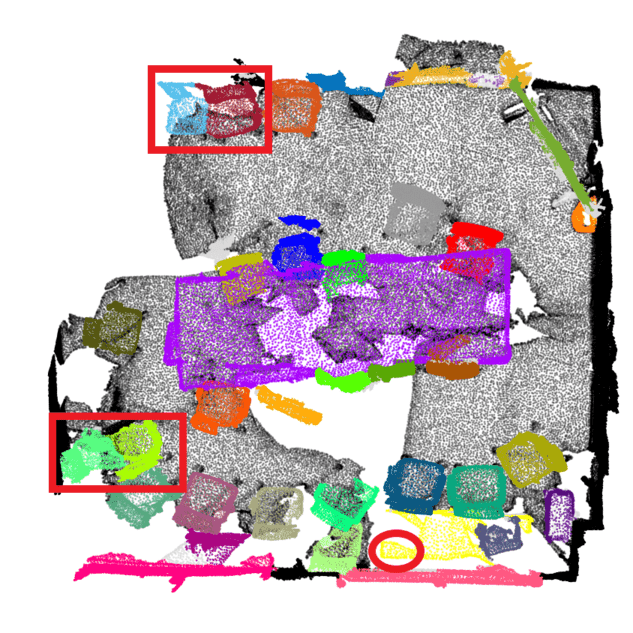}} 
		\centerline{\includegraphics[width=\textwidth]{ 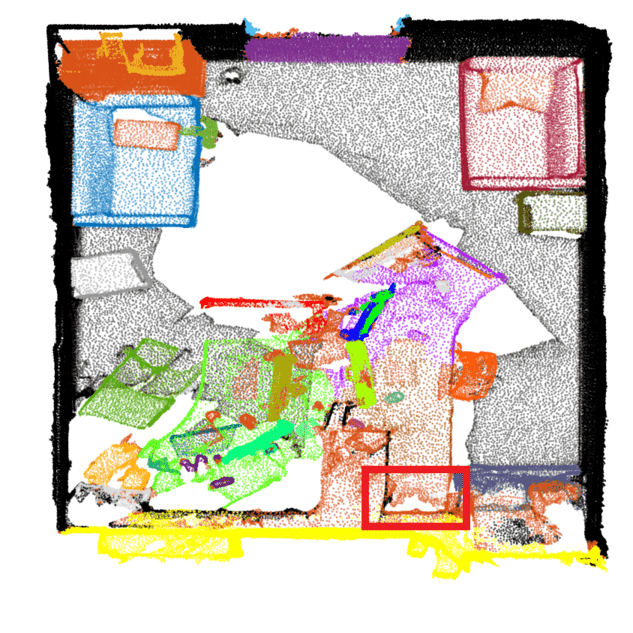}}
		\centerline{Ours }
	\end{minipage}
	\caption{Qualitative Comparison with SOTAs on ScanNetV2. }
	\label{fig:com_pic_sota}
\end{figure*}

\noindent\textbf{Result on ScanNetV2.} Tab.~\ref{tab:com_t} shows the $mAP$ results of PBNet and SOTAs on the hidden test set of ScanNetV2 benchmark. PBNet ranks the first on $mAP$  metric of ScanNetV2 3D instance segmentation challenge, on January 2023. Specifically, PBNet achieves the best performance in 10 out of 18 classes. Following previous work~\cite{wu2022dknet, vu2022softgroup}, we also report the mask segmentation and the detection box results on ScanNetV2 validation set in Tab.~\ref{tab:com_v}. For the mask segmentation, PBNet again shows relative 5.4\%, and 4.3\% improvements on $mAP$ and $AP_{50}$ respectively. On the other hand, our method also gets the best results on Box $AP_{50}$ for the detection task.

\noindent\textbf{Result on S3DIS}.
Following SOTAs, we report the results of Area 5 and 6-fold cross-validation on the S3DIS dataset  in Tab.~\ref{tab:res_s3dis}. For the 6-fold cross-validation, we report the average results. As observed, our approach is still ahead of the other methods on the major metrics $mAP$ and $AP_{50}$. When evaluated on Area 5,  our method shows  the best result on three over four metrics, i.e., $mAP$, $AP_{50}$, and $mPre_{50}$. As for the metric of $mRec_{50}$, our method is inferior to Softgroup~\cite{vu2022softgroup} but still competitive, which ranks the second among all the methods. In the results of 6-fold cross-validation, our model attains the best $mAP$ and $AP_{50}$, and it ranks the second and third respectively on  $mRec_{50}$ and $mPre_{50}$. In short, our model demonstrates the overall best performance on both Area 5 and 6-fold cross-validation. 

\noindent\textbf{Qualitative Comparison}. We also provide qualitative comparisons based on ScanNetV2 (see Fig.~\ref{fig:com_pic_sota}). Clearly, our method exhibits visually better performance than the other SOTAs. More visualization results on both ScanNetV2 and S3DIS are provided in the supplementary.

%========================================================Ablation===================================================================================
\subsection{Ablation Study and Analysis}
\begin{table}[b]
	%表格剧中
	\centering
	\resizebox{0.40\textwidth}{!}{
	\begin{tabular}{ccc|ccc}
		\bottomrule
		Distance & Binary  & Instance   & $mAP$   & $AP_{50}$ &$AP_{25}$ \\
		Clustering &Clustering &Refine & & &                                    \\\hline
	    $\sqrt{ }$& &  & 48.9&66.9 &77.9 \\
		 & $\sqrt{ }$ & &50.4 & 68.3&78.6               \\
		 & $\sqrt{ }$   & $\sqrt{ }$  & \textbf{54.3} &\textbf{70.5} & \textbf{78.9}              \\
     \bottomrule
	\end{tabular}
	}
	\caption{Ablation Study on Network Module.}
	%设置label
	\label{tab:abl_mod}
\end{table}
To verify the effectiveness of our method, in this section, we conduct ablation experiments and parameter sensitivity analysis on the ScanNetV2 validation set. First, we verify two main modules in the network architecture~(see Fig.~\ref{fig:net}): Binary Clustering (b) and Instance Refine (c). As shown in Tab.~\ref{tab:abl_mod}, PBNet achieves significant improvements compared to the baseline. Comparing distance clustering, our binary clustering attains improvements on all three metric:$mAP$, $AP_{50}$ and $AP_{25}$. Meanwhile, our refinement method based on local scenes also plays a vital role in  improving  performance. Remarkably, it manages to improve 4.1\% w.r.t. $mAP$ when the refinement is applied on top of binary clustering, which clearly demonstrates its effectiveness. 

\begin{figure}[t]
	\centering 
	\begin{minipage}{0.22\linewidth}
		%		\begin{mdframed} 
		\centerline{\includegraphics[width=\textwidth]{ 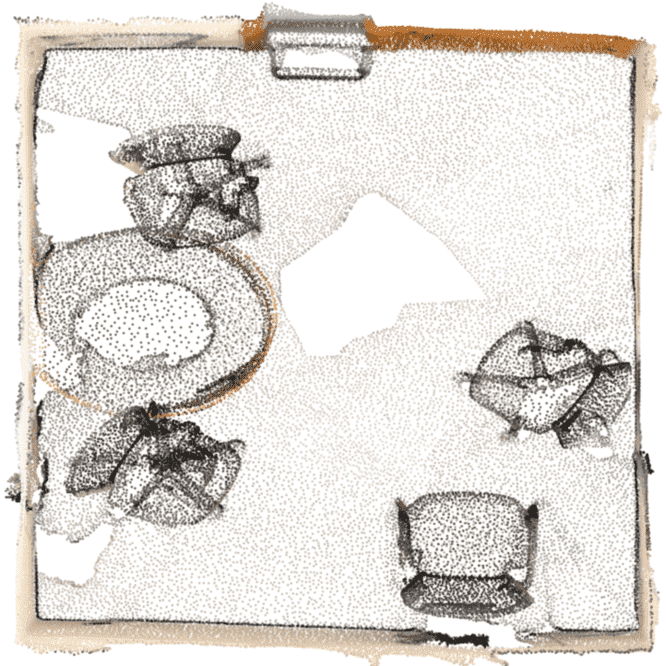}}
		\centerline{Input}
		%		\end{mdframed}	
	\end{minipage}
	\begin{minipage}{0.22\linewidth} 
		%		\begin{mdframed} 
		\centerline{\includegraphics[width=\textwidth]{ 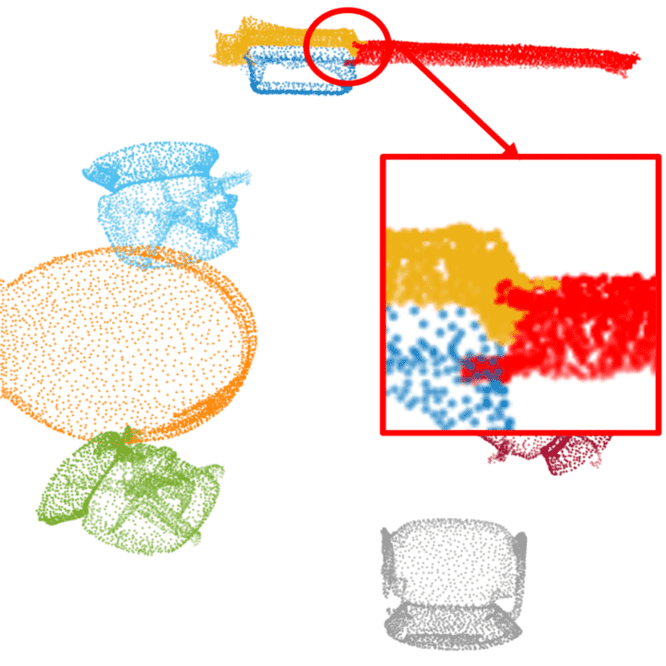}}
		\centerline{Instance GT}
		%		\end{mdframed}	
	\end{minipage}
	\begin{minipage}{0.22\linewidth}
		%\vspace{3pt}  是调图片之间的间隔 
		%		\begin{mdframed} 
		\centerline{\includegraphics[width=\textwidth]{ 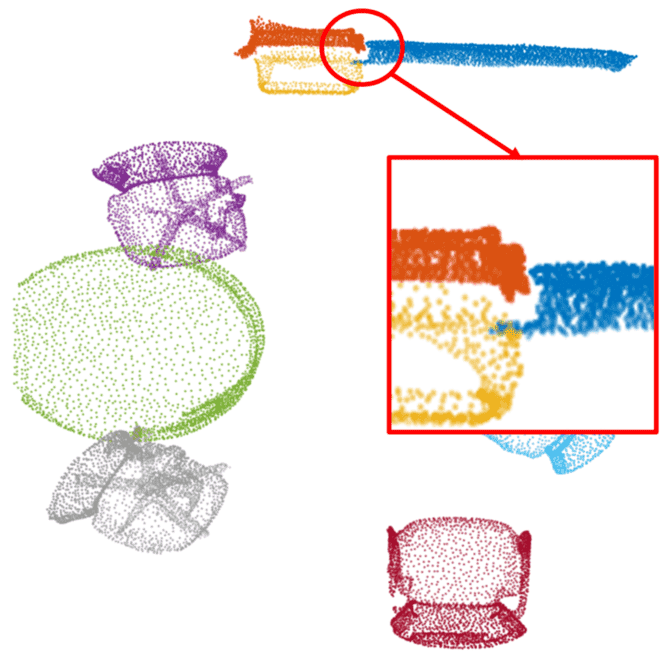}}
		\centerline{$w/o$ Voting}
		%		\end{mdframed}	
		
	\end{minipage}
	\begin{minipage}{0.22\linewidth}
		%\vspace{3pt}  是调图片之间的间隔
		%		\begin{mdframed} 
		\centerline{\includegraphics[width=\textwidth]{ 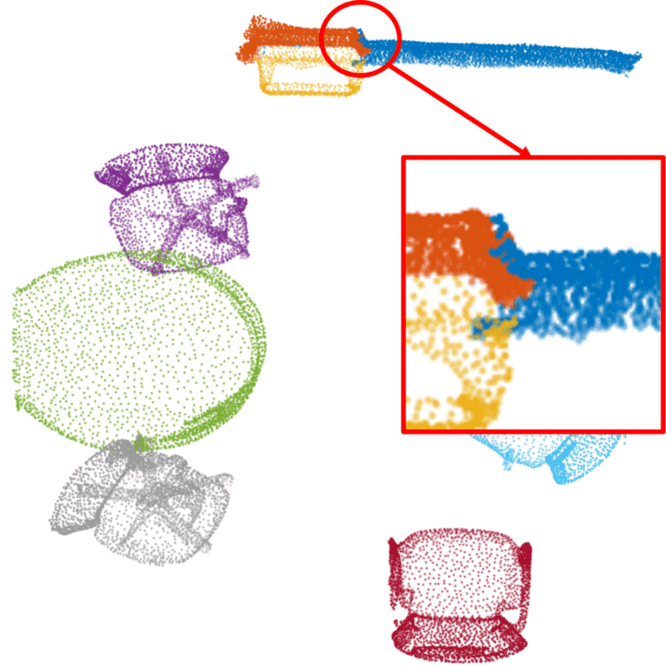}}
		%		\end{mdframed}	
		\centerline{$w$ Voting}
	\end{minipage}
	\caption{Ablation Study on Voting LPs. Red boxes highlight the difference between results with ($w$) and without ($w/o$) voting LPs.}
	\label{fig:res_abl_vote}
\end{figure}

\noindent\textbf{Ablation on Binary Clustering.}
We conduct further analysis on the effectiveness of binary clustering. Specifically, binary clustering includes Group HPs and Voting LPs. Tab.~\ref{tab:abl_bin} analyzes the effectiveness of each part. In the part without LPs, we take LPs as background points. Evidently, a combination of both parts could lead to the best results.  Qualitative comparison between with($w$) and without($w/o$) voting LPs in Fig.~\ref{fig:res_abl_vote} also validates this point. 
\begin{table}[ht]
	%表格剧中
	\centering
	\resizebox{0.32\textwidth}{!}{
	\begin{tabular}{cc|ccc}
		\bottomrule
	Group HPs& Vote LPs    & $mAP$   & $AP_{50}$ &$AP_{25}$ \\\hline
	 $\sqrt{ }$ &          &52.9      &68.7      &78.4               \\
	$\sqrt{ }$ & $\sqrt{ }$ & \textbf{54.3}    &\textbf{70.5}        & \textbf{78.9}              \\
     \bottomrule
	\end{tabular}
	}
	\caption{Ablation Study on Binary Clustering.}
	%设置label
	\label{tab:abl_bin}
\end{table}

In addition, we examine if our binary clustering idea could work as a plug-in to improve other mainstream baselines. To this end, we take the baselines PointGroup~\cite{jiang2020pointgroup} and HAIS~\cite{chen2021hierarchical} as two typical examples where we  simply replace the traditional distance clustering with our binary clustering. To be fair, we directly use their published pre-trained model for validation. We report the results in  Tab.~\ref{tab:add_bc}. As clearly observed, our binary clustering leads to substantial improvements as opposed to distance clustering, verifying the advantages of our proposed method.
\begin{table}[ht]
	%调整行距
% 	\renewcommand{\arraystretch}{1.3}
	%表格剧中
	\centering
	\resizebox{0.39\textwidth}{!}{
	\begin{tabular}{c|ccc}
	\bottomrule
	Baseline Model	   & $mAP$   & $AP_{50}$ &$AP_{25}$ \\
	\hline
	PointGroup~\cite{jiang2020pointgroup} & 35.5(+2.0\%) & 58.4(+2.6\%) &72.3(+1.4\%) \\
	HAIS~\cite{chen2021hierarchical}  & 44.7(+2.8\%) &65.7(+2.5\%) & 76.0(+0.5\%) \\
	\bottomrule
	\end{tabular}
	}
	%设置表标题在下面
	\caption{Our binary clustering leads to consistent improvement  by simply replacing distance clustering on various baselines.}
	%设置label
	\label{tab:add_bc}
\end{table}

\noindent\textbf{Ablation on Instance Refine.} 
In Tab.~\ref{tab:abl_loc}, we report the ablation experiment results of instance refine. Notably, the proposed mask loss  boosts both the $mAP$ and $AP_{50}$ metric. Combined with the local scene mechanism, instance refine increases the performance w.r.t. all three metrics. Particularly, it
improves by a relative 7.7\% and 3.2\% on $mAP$ and $AP_{50}$ against the baseline respectively.
\begin{table}[h]
	%表格剧中
	\centering
	\resizebox{0.40\textwidth}{!}{
	\begin{tabular}{ccc|ccc}
		\bottomrule
		 Baseline&Local Scene & Mask Loss & $mAP$   & $AP_{50}$ &$AP_{25}$ \\\hline
	    $\sqrt{ }$&           &        &50.4 & 68.3&78.6\\
		          &           &$\sqrt{ }$&53.0 &68.8 & 78.5              \\
		          &$\sqrt{ }$&$\sqrt{ }$&\textbf{54.3}&\textbf{70.5}&\textbf{78.9}              \\
     \bottomrule
	\end{tabular}
	}
	\caption{Ablation Study on Instance Refine.}
	%设置label
	\label{tab:abl_loc}
\end{table}

\subsection{Efficiency}
We examine the efficiency of our PBNet in this section. A single RTX 3090 is adopted to conduct this experiment on the ScanNetV2 validation set. In detail, we report the average inference time for each component of our network architecture in Tab.~\ref{tab:inf}. The baseline includes backbone, 3D convolution, MLP, and data conversion.

\begin{table}[ht]
	%表格剧中
	\centering
	\resizebox{0.47\textwidth}{!}{
	\begin{tabular}{ccccc|c}
		\bottomrule
	Baseline &	Group HPs & Vote LPs  & Local Scene & Post-process & Infer. Time(ms)    \\
	$\sqrt{ }$& &  &  &   & 190.8  \\
	$\sqrt{ }$&$\sqrt{ }$&  &  &   & 322.9  \\
    $\sqrt{ }$&$\sqrt{ }$& $\sqrt{ }$&      & & 339.5              \\
    $\sqrt{ }$&$\sqrt{ }$& $\sqrt{ }$   & $\sqrt{ }$  &  & 402.0         \\
	$\sqrt{ }$&$\sqrt{ }$& $\sqrt{ }$   & $\sqrt{ }$  & $\sqrt{ }$ & 420.8              \\
     \bottomrule
	\end{tabular}
	}
	\caption{Average Inference Time(per scene).}
	%设置label
	\label{tab:inf}
\end{table}
As shown in Tab.~\ref{tab:inf_com}, our PBNet takes an average of 420ms for each 3D scene inference on a single RTX 3090, which is still efficient in practice.  Furthermore, HAIS~\cite{chen2021hierarchical} is currently the fastest inference method for 3D instance segmentation. In contrast, our PBNet only introduces limited latency (150-250 ms) but achieves a significant $mAP$ improvement. Compared with another fast model DKNet~\cite{wu2022dknet}, our method is slightly slower with a limited latency (about 63 ms). Overall, our algorithm is still reasonably efficient though it is slower than HAIS and DKNet. Given the significant mAP improvement, we believe it is a worthwhile trade-off and we will leave the exploration of speeding up our algorithm as future work.

\begin{table}[h]
	%表格剧中
	\centering
	\resizebox{0.35\textwidth}{!}{
	\begin{tabular}{c|ccc|c}
	\bottomrule
	Methods &	$mAP$ & $AP_{50}$  & $AP_{25}$ & Infer. Time(ms)    \\\hline
	HAIS~\cite{chen2021hierarchical}   &43.5 &64.1 &75.6& \textbf{206.0} \\
	DKNet~\cite{wu2022dknet}     &51.5 &67.0 &77.0& 357.5 \\
	Ours      &\textbf{54.3}&\textbf{70.5}&\textbf{78.9}& 420.8 \\ 
    \bottomrule
	\end{tabular}
	}
	\caption{Average Inference Time Comparison (per scene).}
	%设置label
	\label{tab:inf_com}
\end{table}

\begin{figure}[h]
	\centering 
	\begin{minipage}{0.44\linewidth} 
		%		\begin{mdframed} 
		\centerline{\includegraphics[width=\textwidth]{ 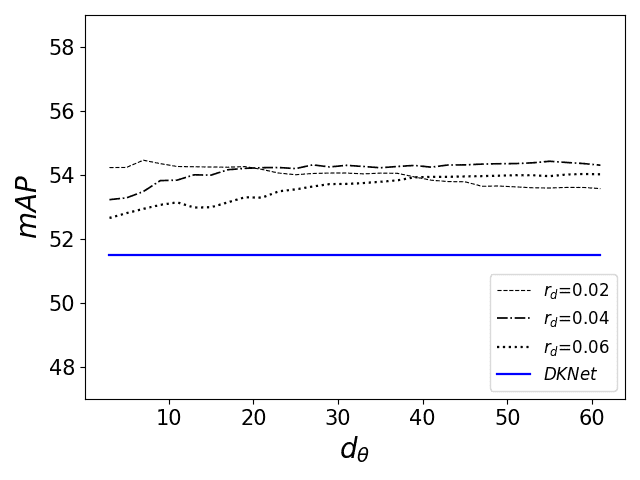}}
		%		\end{mdframed}	
	\end{minipage}
	\hspace{0.06\linewidth}
	\begin{minipage}{0.44\linewidth} 
		%		\begin{mdframed} 
		\centerline{\includegraphics[width=\textwidth]{ 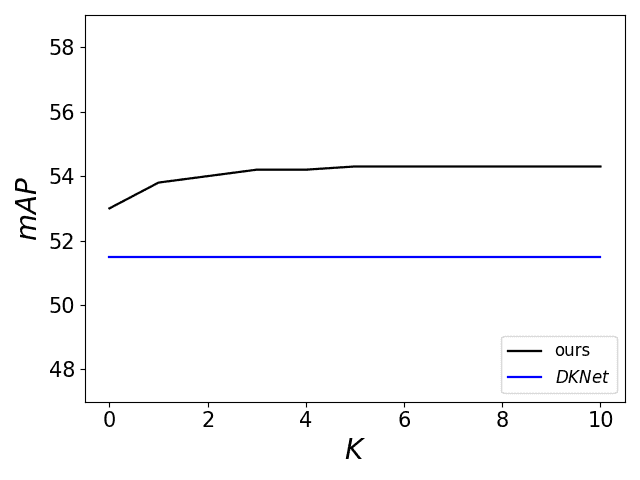}}
		%		\end{mdframed}	
	\end{minipage}
	\caption{Parameter Sensitivity Analysis.}
	\label{fig:sen}
\end{figure}

%========================================================Parameter===================================================================================
\noindent\textbf{Parameter Analysis.} The clustering-based methods all contain fine-tuning parameters. For example, DKNet~\cite{wu2022dknet} includes three parameters: $r_d$,  $\alpha$, and $T_\theta$, where $\alpha$ is the formula coefficient and  $T_\theta$ is the normalized centroid score threshold; RPGN~\cite{dong2022learning} has five parameters. In comparison, our method needs three parameters: $r_d$,  $d_{\theta}$, $K$, where $r_d$ and  $d_{\theta}$ are used in binary clustering, and $K$ is for constructing local scenes. In Fig.~\ref{fig:sen}, we conduct parameter sensitivity analysis on the ScanNetV2 validation set.

Specifically, we plot $mAP$ by  setting $r_{d}$ to 0.02, 0.04 and 0.06, and continuously increasing the density threshold $d_{\theta}$. Obviously, when $r_{d}$ is 0.04,  $mAP$ appears stable especially when $d_{\theta}$ is greater than 20. In our experiments, we thus set $r_{d}$ and $d_{\theta}$ to 0.04 and 30 respectively. We also  evaluate $mAP$ vs. $K$. When $K$ is greater than 6, $mAP$ stabilizes and does not change. Hence, we set $K$ to 6. Overall, the number of hyperparameters is three in our method, which is parred to or fewer than that of SOTAs. All these parameters appear less sensitive as  observed empirically.

%========================================================Conclusion====================================================================================
\section{Conclusion}
We propose a novel divide and conquer strategy for 3D point cloud instance segmentation with point-wise binarization. Termed as PBNet, our end-to-end network makes a first attempt to divide offset instance points into two categories: high and low density points (HPs vs. LPs). While HPs can be leveraged to separate adjacent objects confidently, LPs can help complete and refine instances via a novel neighbor voting scheme. We have  developed a local scene mechanism to refine instances and suppress over-segmentation. Extensive  experiments on benchmark ScanNetV2 and S3DIS datasets have shown that our model can  overall beat  the existing best models. In the future, we will explore how to speed up our algorithm.
\\

\noindent\textbf{Acknowledgement.} This work was in part supported by the Jiangsu Science and Technology Programme (Natural Science Foundation of Jiangsu Province) under No.BE2020006-4, the Natural Science Foundation of the Jiangsu Higher Education Institutions of China under No.22KJB520039, and the National Natural Science Foundation of China under No.62206225.

{\small
\bibliographystyle{ieee_fullname}
\bibliography{egbib}
}

\end{document}